\documentclass[10pt,twocolumn,letterpaper]{article}

\usepackage[pagenumbers]{iccv} 

\usepackage{times}
\usepackage{epsfig}
\usepackage{graphicx}
\usepackage{amsmath}
\usepackage{amssymb}

\usepackage{booktabs}   
\usepackage{siunitx}    
\usepackage{multirow}   
\usepackage{amsfonts}   
\usepackage{subcaption} 
\usepackage{placeins} 
\usepackage{adjustbox}
\usepackage{stfloats}

\definecolor{iccvblue}{rgb}{0.21,0.49,0.74}
\usepackage[pagebackref,breaklinks,colorlinks,allcolors=iccvblue]{hyperref}

\def\paperID{*****} 
\def\confName{ICCV}
\def\confYear{2025}

\title{VeS: Teaching Pixels to Listen Without Supervision}

\author{Sajay Raj\\
Indian Institute of Technology, Madras\\
{\tt\small sajayraj08@gmail.com}
}

\begin{document}
\maketitle

\begin{abstract}
Recent dense audio--visual (AV) models achieve impressive retrieval and emergent localization \cite{hamilton2024separating,radford2021learning}, but almost all evidence comes from English‑centric, caption‑rich web video. It is unclear whether these objectives \emph{survive} in low‑resource, code‑switched, and noisy multilingual settings that typify developing regions. We show they do—and that the \textbf{choice of aggregation function} becomes even more critical. Using a multilingual subset of \textbf{Project Vaani} \cite{vaani2024site,vaani2024hf} spanning dozens of Indian languages and dialectal variants, we compare three contrastive objectives: (i) a global mean‑pooled loss (CLIP‑style \cite{radford2021learning}), (ii) a dense max--mean token matcher (DenseAV‑style \cite{hamilton2024separating}), and (iii) a simple hybrid (motivated by frozen‑vision alignment strategies \cite{zhai2022lit,jose2025dinov2text}). The dense objective delivers a \textbf{+59\% relative R@1} (Audio$\to$Visual) improvement over global pooling and substantially lower mean/median ranks, while consistently producing \emph{sharp} zero‑shot localization heatmaps of spoken objects—despite keeping the vision backbone \emph{entirely frozen} (no LoRA / partial fine-tuning). Our results demonstrate that dense token routing is \emph{not} a luxury of high‑resource English corpora; it is \emph{more} decisive when annotations and acoustic cleanliness are scarce. We release the codebase and trained models\footnote{Code: \url{https://github.com/SajayR/VeS}, Models: \url{https://huggingface.co/SajayR/VeS}}.
\end{abstract}
\section{Introduction}
\label{sec:intro}
Self-supervised audio--visual (AV) learning has advanced rapidly, yielding impressive retrieval and emergent localization capabilities \cite{hamilton2024separating,radford2021learning}, yet most progress relies on abundant, relatively clean, predominantly English web data, leaving a gap for linguistically dense but annotation‑scarce contexts such as India \cite{vaani2024site,vaani2024hf}.

We probe which contrastive \emph{aggregation} strategy is most effective for noisy, multilingual speech--image alignment when no text or spatial supervision is available. Using the large-scale \textbf{Project Vaani} dataset \cite{vaani2024site,vaani2024hf}, we compare three approaches under identical backbones: (1) a global CLIP-style mean-pooled loss \cite{radford2021learning}, (2) a dense max--mean token-matching loss \cite{hamilton2024separating}, and (3) a trival hybrid of the two.

\textbf{Key finding.} The aggregation function is critical. The dense objective outperforms global pooling (\textbf{+59\% relative R@1 A$\to$V}) and consistently produces sharp zero-shot localization of spoken objects. This indicates dense token routing is not a luxury of high-resource corpora but becomes \emph{more} decisive when data quality and annotations are scarce.

All experiments run on a \textbf{single 24\,GB consumer GPU}, enabled by freezing the vision backbone \cite{oquab2023dinov2,zhai2022lit} and lightweight adapters \cite{jose2025dinov2text}, underscoring accessibility. We release code and models to foster inclusive AV research.

\subsection{Existing Work}
\label{sec:existing-work}

\textbf{Audio--visual contrastive learning.}
Early work on cross-modal retrieval adopted global pooling objectives from image--text pre-training. CLIP's global InfoNCE loss~\cite{radford2021learning} delivers strong zero-shot transfer but discards spatial/temporal structure critical for localization. LiT showed that competitive transfer is possible even when the vision backbone is \emph{frozen} and only a projection head is tuned~\cite{zhai2022lit}, an idea we adopt to keep training on a single 24\,GB GPU.

\textbf{Late-interaction / token-level objectives.}
DenseAV introduced a bidirectional max--mean aggregation that forces each audio token to align with at least one visual patch, yielding sharp emergent heatmaps~\cite{hamilton2024separating}. FILIP utilized the same late-interaction idea to image--text by averaging the two directional maxima to capture fine-grained correspondences without cross-attention~\cite{yao2022filip}. Our dense loss matches DenseAV's audio$\rightarrow$visual direction; our experiments showed degraded localization in the multilingual regime with FILIP-style bi-directional objective.

\textbf{Frozen vision backbones with lightweight adapters.}
Locked-image tuning (LiT)~\cite{zhai2022lit} and, more recently, DINOv2 Meets Text~\cite{jose2025dinov2text} show that keeping strong self-supervised ViTs frozen while attaching tiny adapters preserves performance and slashes memory. We reuse this strategy, caching DINOv2~\cite{oquab2023dinov2} patch tokens once and training only the audio branch plus two vision adapters.

\textbf{Multilingual, low-resource speech.}
Prior contrastive AV work largely focuses on English captions (e.g., YouTube/HowTo100M), leaving open whether dense objectives survive noisy, code-switched speech. Large self-supervised speech encoders such as wav2vec~2.0~\cite{baevski2020wav2vec2} and DistilHuBERT~\cite{chang2021distilhubert} shrink the gap for low-resource languages but are rarely coupled with vision data at scale. Project Vaani~\cite{vaani2024site,vaani2024hf} supplies the missing multilingual corpus; to our knowledge, we are the first to benchmark dense, global, and hybrid AV losses on this data.

\textbf{Summary.}
Our study sits at the intersection of these threads: we fuse a frozen, patch-rich ViT with a multilingual speech encoder and systematically compare global, dense, and hybrid contrastive objectives. Results (\S3) confirm that late-interaction routing is \emph{more} decisive---not less---in noisy, annotation-scarce settings.

\section{Methodology}
\label{sec:method}

Our approach is to train an audio-visual alignment model using a frozen image encoder and a trainable audio encoder. We freeze\footnote{Freezing also permits one‑time offline extraction and caching of patch tokens, allowing all loss variants to reuse identical vision features and run on a single 24\,GB GPU by offloading the vision encoder during training.}
a pre-trained \textbf{DINOv2}~\cite{oquab2023dinov2} ViT as our vision backbone due to its strong performance and rich patch-level features. The audio is processed by a trainable transformer-based model. The core of our investigation lies in the formulation of the similarity score used within a contrastive loss framework.

\subsection{Model Architecture and Efficiency Choices}
\label{sec:arch}

\paragraph{Overview.}
Our model factorizes into an \textbf{audio branch}, a \textbf{frozen vision backbone}, light \textbf{vision adapters}, and a \textbf{dense similarity module}. Given raw 16\,kHz audio and a single video frame $I$, we produce $\ell_2$-normalized token matrices $A\in\mathbb{R}^{N_a\times D}$ and $V\in\mathbb{R}^{N_v\times D}$ which feed the aggregation functions of Section~\ref{sec:method}.

\paragraph{Vision side (frozen + cached).}
We use a DINOv2-Large encoder to obtain patch + class tokens. The patch tokens are extracted once and cached (BF16), while the [CLS] token is discarded. Two lightweight transformer encoder layers (\emph{adapter}) operate on cached tokens before a linear--LayerNorm--linear projection to dimension $D=256$. 

\paragraph{Audio side (temporal downsampling).}
The audio branch is a HuBERT-based transformer (DistilHuBERT) producing a sequence of hidden states. We apply a stride‑2 average pooling over time (after the last hidden layer) reducing token count from $N_a^{(0)}$ to $N_a = N_a^{(0)}/2$. A binary mask $M$ (silence / padding) is downsampled via pooling and thresholding. This single pooling:
(i) lowers the memory/time of the dense similarity volume ($O(N_a N_v)$) roughly by factor~2,  
(ii) smooths very short transients that otherwise create spurious heatmaps

\paragraph{Projection \& normalization.}
Both modalities pass through small projection heads (linear / LayerNorm / linear) to a shared embedding dimension, then are $\ell_2$-normalized. No cross-attention or fusion layers are introduced; all cross-modal coupling happens only inside the aggregation operator.

\subsection{Loss Function Investigation}
Our investigation compares three contrastive loss formulations, each built from a different method of aggregating token-level similarities into a clip-level score. Let $A^{(b)} \in \mathbb{R}^{N_a^{(b)} \times D}$ and $V^{(b)} \in \mathbb{R}^{N_v \times D}$ be the $\ell_2$-normalized audio token and visual patch feature matrices for the $b$-th sample in a batch of size $B$.

\subsubsection{Dense Audio-Visual Similarity (Dense Loss)}
First, for every pair of samples $(b, b')$ in the batch, we compute a dense, token-level similarity matrix by taking the inner product between all audio tokens from sample $b$ and all visual patch tokens from sample $b'$:
$S^{(b,b')} = A^{(b)} (V^{(b')})^T \in \mathbb R^{N_a^{(b)}\times N_v}$.

Inspired by \cite{hamilton2024separating}, we compute a final clip-level similarity by bidirectionally aggregating these token similarities. Let $M^{(b)} \in \{0,1\}^{N_a^{(b)}}$ be a mask indicating non-silent audio tokens and $\varepsilon=10^{-6}$ for numerical safety. The audio-to-visual similarity $\Phi$ aggregates by taking the \textbf{max} over visual patches for each audio token and then the \textbf{mean} across time:
\begin{equation}
\Phi(S, M) = \frac{\displaystyle \sum_{t=1}^{N_a^{(b)}} M_t \max_{p} S_{t,p}}{\displaystyle \sum_{t=1}^{N_a^{(b)}} M_t + \varepsilon}.
\end{equation}
Conversely, inspired by \cite{yao2022filip}, we evaluated a visual-to-audio similarity $\Psi$ aggregation by taking \textbf{max} over time for each visual patch and then the \textbf{mean} across patches.:
\begin{equation}
\Psi(S) = \frac{1}{N_v}\sum_{p=1}^{N_v} \max_{t} S_{t,p}.
\end{equation}
This was found to be detrimental for feature localization and provided minor improvements to retrieval scores and hence was discarded.

The final dense clip-level similarity for the pair $(b,b')$ is the hence the audio-to-visual similarity:
\begin{equation}
C_{b,b'}^{\text{dense}} =  \Phi(S^{(b,b')}, M^{(b)}) .
\label{eq:clip_sim_dense}
\end{equation}
This max-pooling mechanism forces the model to find at least one strong alignment for each token, encouraging sparse and meaningful local feature alignment, which is crucial for localization.

\subsubsection{Global Mean-Pooled Similarity (Global Loss)}
To compute a global similarity, we first discard all spatial and temporal information by aggregating the tokens for each modality into a single global feature vector. The global audio feature $\bar{a}^{(b)}$ is the masked average of its tokens, and the global visual feature $\bar{v}^{(b)}$ is the mean of its patches:
\begin{align}
\bar{a}^{(b)} &= \frac{\sum_{t=1}^{N_a^{(b)}} M_t^{(b)} a_t^{(b)}}{\sum_{t=1}^{N_a^{(b)}} M_t^{(b)} + \varepsilon}, \\
\bar{v}^{(b)} &= \frac{1}{N_v} \sum_{p=1}^{N_v} v_p^{(b)}.
\end{align}
The clip-level similarity for the pair $(b,b')$ is then the cosine similarity between their global feature vectors:
\begin{equation}
C_{b,b'}^{\text{global}} = \langle \bar{a}^{(b)}, \bar{v}^{(b')} \rangle.
\label{eq:clip_sim_global}
\end{equation}

\subsubsection{Hybrid Loss.}
We also evaluate a simple hybrid contrastive loss that linearly combines the dense and global variants:
\begin{equation}
\mathcal{L}_{\text{hyb}} = \lambda \mathcal{L}_{\text{dense}} + (1-\lambda)\mathcal{L}_{\text{global}},
\label{eq:hybrid}
\end{equation}
with a fixed $\lambda=0.7$ (For higher localization objective maximization) 

\subsubsection{Contrastive Objective}
Let $C_{b,b'}$ be the clip-level similarity matrix from either the dense (Eq.~\ref{eq:clip_sim_dense}) or global (Eq.~\ref{eq:clip_sim_global}) method. We use a standard symmetric InfoNCE loss with a learnable temperature parameter $\tau$:
\begin{equation}
\mathcal{L}_{\text{contr}} = -\frac{1}{2B} \sum_{b=1}^{B} \left( \log \frac{e^{\tau C_{b,b}}}{\sum_{k=1}^B e^{\tau C_{b,k}}} + \log \frac{e^{\tau C_{b,b}}}{\sum_{k=1}^B e^{\tau C_{k,b}}} \right).
\label{eq:contrastive}
\end{equation}

\section{Dataset and Training Setup}
\label{sec:data}

\paragraph{Corpus.}
We use a large multilingual subset of Project Vaani (CC-BY-4.0) \cite{vaani2024site,vaani2024hf} comprising 4.83M speech segments with 83 language / dialect labels (e.g., Hindi, Bengali, Tamil, Telugu, Marathi, Bhojpuri, Maithili, Kannada, Nepali). No textual supervision (transcripts or captions) or spatial annotations are used at any stage; Code-switching and background noise are common.

\paragraph{Splits.}
We hold out 5{,}000 audio--visual pairs for validation; the rest form the training set. The validation split is stratified across high-, mid-, and low-frequency labels to remain challenging and prevent dominance by a single high-resource language.

\paragraph{Preprocessing.}
Audio is resampled (if needed) and peak-normalized, then trimmed or zero-padded to a maximum of $5$\,s (binary mask $M$ marks valid tokens). The images are resized to $224{\times}224$, and encoded into patch tokens. All token embeddings are $\ell_2$-normalized before similarity.

\paragraph{Optimization.}
We use 8-bit Adam (bitsandbytes) with mixed-precision (BF16 autocast). The base learning rate is $3{\times}10^{-4}$ with cosine decay (with warmup). Gradient accumulation emulates a larger effective batch under memory limits; although accumulated micro-batches do not create true in-batch negatives, all variants share identical accumulation and optimizer settings.

\paragraph{Compute.}
All experiments run on a single 24\,GB RTX\,4090. Freezing and BF16 caching of vision patch tokens shifts per-step cost almost entirely to the audio branch plus the $O(N_aN_v)$ similarity aggregation, enabling reproducibility without multi-GPU infrastructure.

\paragraph{Metrics.}
We report Audio$\to$Visual and Visual$\to$Audio Recall@K ($K\!\in\!\{1,5,10,50\}$), mean rank, and median rank on the 5k split; a uniform random baseline is provided. Localization is assessed qualitatively via per-token heatmaps (Section~\ref{sec:experiments}).

\section{Experiments}
\label{sec:experiments}

\begin{table*}[t]
\centering
\caption{Retrieval performance on the 5k Vaani validation set. Dense loss dominates across both Audio-to-Visual (A2V) and Visual-to-Audio (V2A) retrieval.}
\label{tab:retrieval_full}
\sisetup{
    table-format=2.2,
    round-mode=places,
    round-precision=2,
    table-auto-round,
    detect-weight=true,
    detect-family=true
}
\begin{tabular*}{\textwidth}{@{\extracolsep{\fill}} l l
    S[table-format=2.2]
    S[table-format=2.2]
    S[table-format=2.2]
    S[table-format=2.2]
    S[table-format=4.1]
    S[table-format=4.1]
    c @{}}
\toprule
\textbf{Loss Type} & \textbf{Direction} & {\textbf{R@1 (\%)}} & {\textbf{R@5 (\%)}} & {\textbf{R@10 (\%)}} & {\textbf{R@50 (\%)}} & {\textbf{Mean Rank}} & {\textbf{Median Rank}} \\
\midrule
\multirow{2}{*}{\textbf{Dense}} 
 & Audio $\to$ Visual & \bfseries 9.90 & \bfseries 24.06 & \bfseries 32.54 & \bfseries 55.36 & \bfseries 266.0 & \bfseries 35.0   \\
\cmidrule(l){2-8}
 & Visual $\to$ Audio & \bfseries 8.50 & \bfseries 21.18 & \bfseries 29.66 & \bfseries 53.98 & \bfseries 252.4 & \bfseries 39.0 & \\
\midrule
\multirow{2}{*}{Global}
 & Audio $\to$ Visual & 6.22 & 16.52 & 23.88 & 47.52 & 339.8 & 58.0 \\
\cmidrule(l){2-8}
 & Visual $\to$ Audio & 6.38 & 16.54 & 24.08 & 47.24 & 341.8 & 59.5 & \\
\midrule
\multirow{2}{*}{Hybrid}
 & Audio $\to$ Visual & 9.00 & 21.86 & 29.32 & 53.18 & 283.0 & 42.0 \\
\cmidrule(l){2-8}
 & Visual $\to$ Audio & 7.46 & 20.32 & 28.44 & 52.28 & 271.2 & 44.0 & \\
\midrule
\multicolumn{2}{l}{Random (Chance)} & 0.02 & 0.10 & 0.20 & 1.00 & 2501.0 & 2501.0 \\
\bottomrule
\end{tabular*}

\vspace{4pt}
\end{table*}

We conduct experiments on a curated subset of the \textbf{Vaani} dataset, paired with relevant images. We evaluate all models on audio-to-visual (A2V) and visual-to-audio (V2A) retrieval.

\subsection{Retrieval and Localization Results}
Our quantitative results are presented in Table~\ref{tab:retrieval_full}. The model trained with the \textbf{Dense Loss} decisively outperforms the \textbf{Global Loss} and \textbf{Hybrid Loss} model across all standard retrieval metrics, including Recall@K (R@K), mean rank, and median rank. For instance, in A2V retrieval, the Dense model achieves a R@1 of \textbf{9.9\%}, compared to \textbf{6.22\%} for the Global model and \textbf{9.0\%}. This indicates that the fine-grained supervision signal from the dense token comparison provides a much stronger learning signal for associating entire clips, even though it operates at the token level.

\begin{figure*}[!t]
  \centering
  \setlength{\abovecaptionskip}{4pt}
  \setlength{\belowcaptionskip}{0pt}

  \begin{adjustbox}{max totalheight=0.975\textheight, keepaspectratio}
    \begin{minipage}{\textwidth}
      \includegraphics[width=\textwidth]{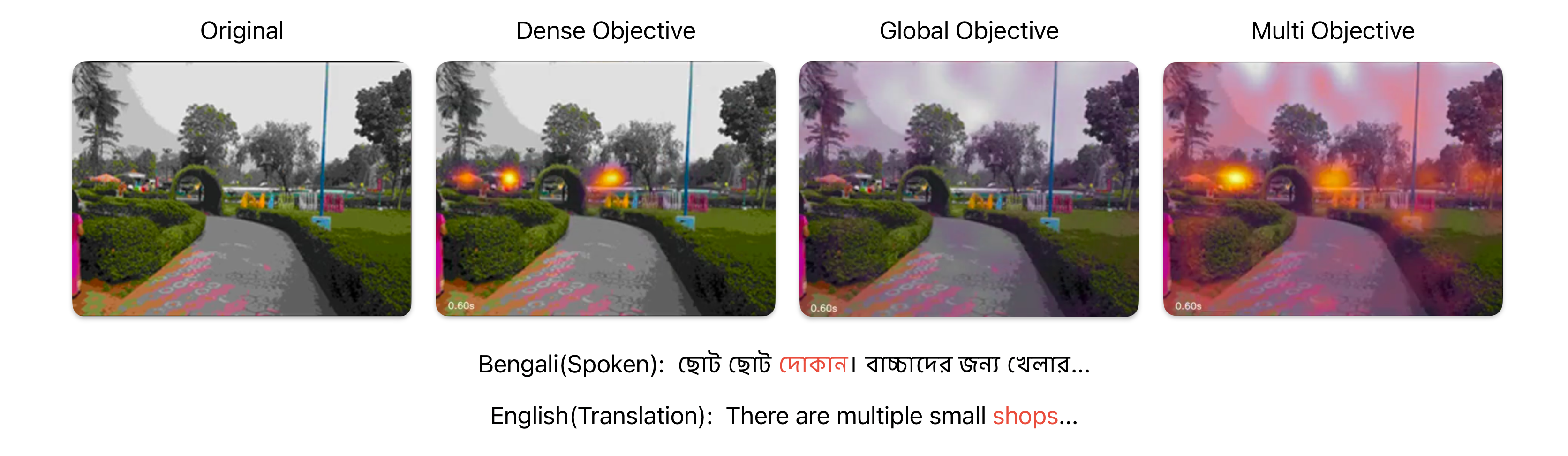}\\[-2pt]
      \includegraphics[width=\textwidth]{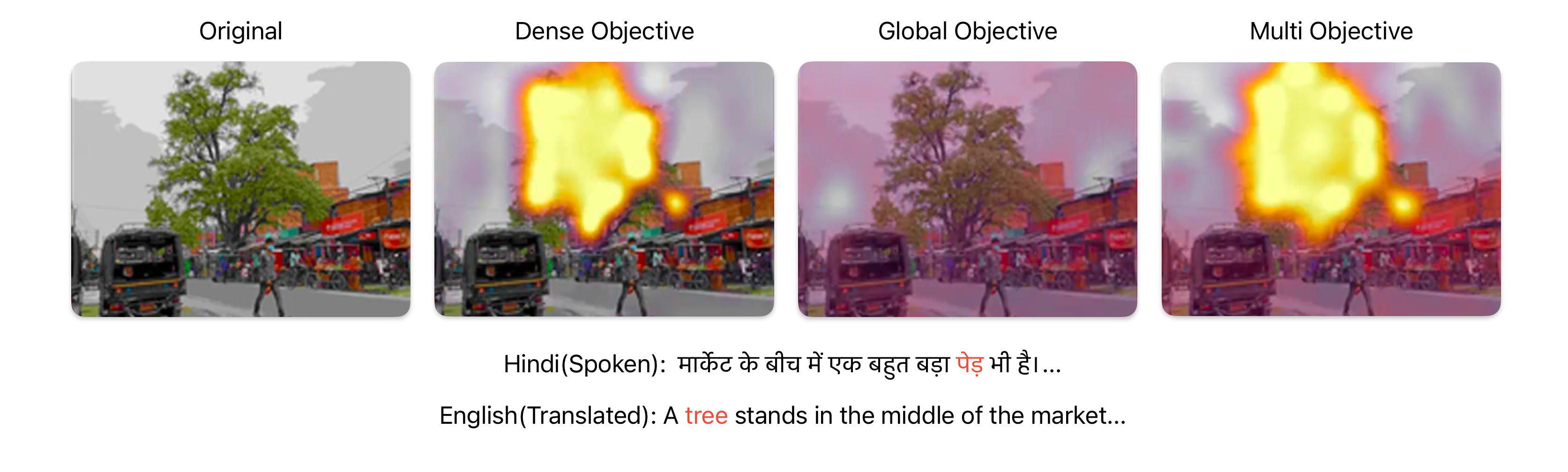}\\[-2pt]
      \includegraphics[width=\textwidth]{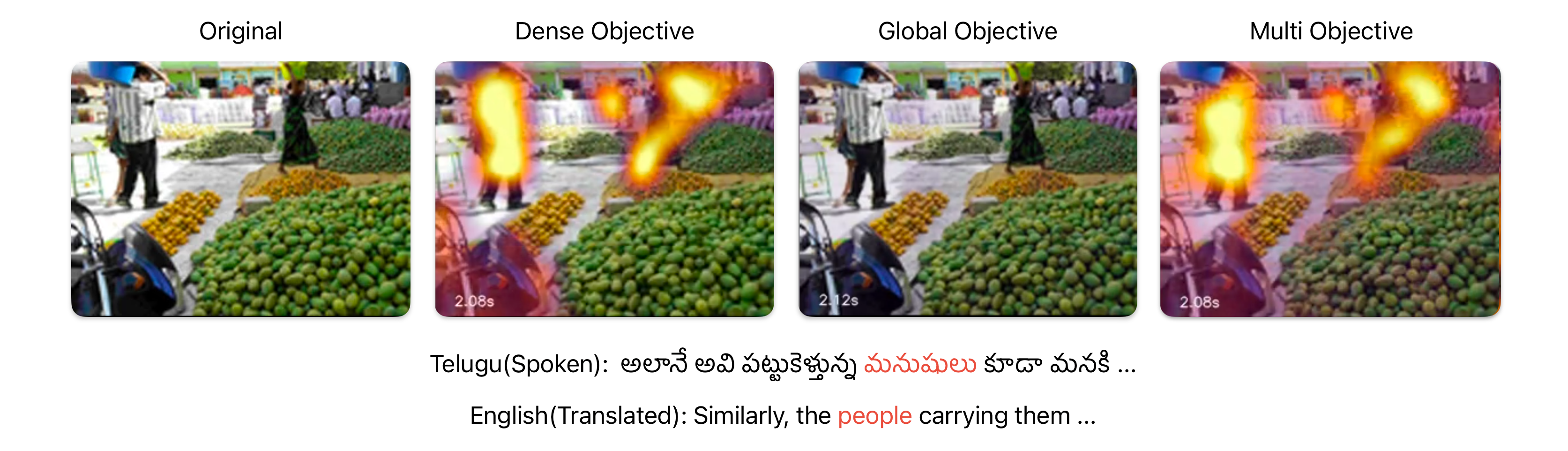}\\[-2pt]
      \includegraphics[width=\textwidth]{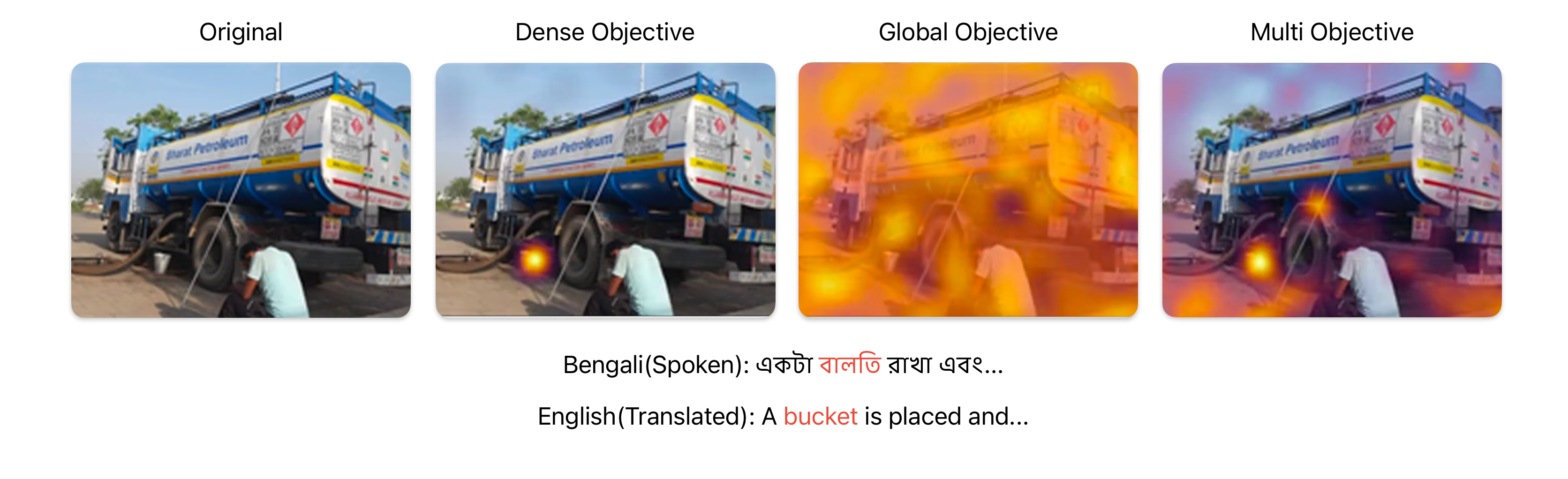}\\[-2pt]
      \includegraphics[width=\textwidth]{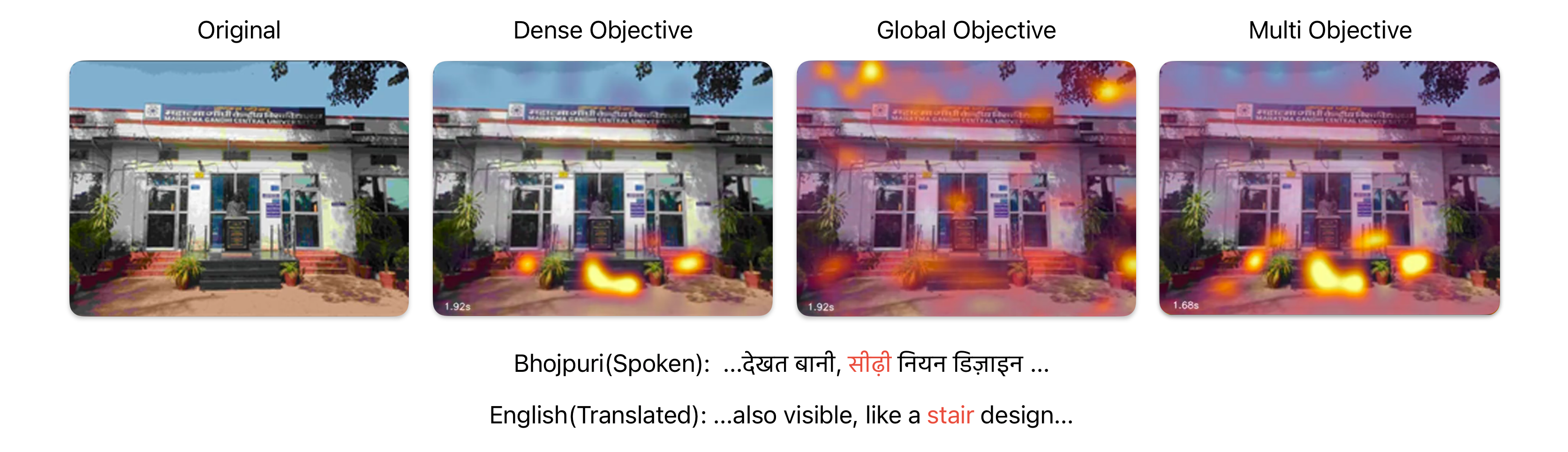}
    \end{minipage}
  \end{adjustbox}

  \caption{Figure 1. Zero-shot localization of spoken objects across multilingual inputs. Dense loss (column 2) produces sharp heatmaps correctly highlighting mentioned objects (shops, tree, people, bucket, door) while global loss (column 3) fails to localize. Multi/Hybrid-objective shows intermediate performance.}
  \label{fig:viz_qual}
\end{figure*}

Most importantly, the choice of loss function has a dramatic impact on the model's ability to perform semantic localization. Figure [~\ref{fig:viz_qual}] shows a comparison of the three losses along with the spoken transcription and a rough translation. The \textbf{Dense Loss} model produces sharp, well-defined heatmaps that accurately locate and highlight the corresponding object when a word is spoken. In contrast, the \textbf{Global Loss} model, which averages away all spatial information before the loss calculation, fails to learn this alignment and produces diffuse, uninformative heatmaps. The Hybrid model offers a compromise, with much better retrieval than the Global model but weaker localization than the pure Dense model. This finding makes intuitive sense.

\section{Conclusion}
\label{sec:conclusion}

In this work, we successfully adapted modern audio-visual learning techniques to the challenging, low-resource domain of Indian languages using the Vaani dataset. We demonstrated that the method of aggregating similarities for contrastive learning is a critical design choice. A \textbf{dense, token-level comparison} is vastly superior to a global comparison, yielding significant gains in both cross-modal retrieval and, crucially, enabling zero-shot semantic localization. Our findings provide a clear path for developing more capable and inclusive AV systems that can serve a global audience, moving beyond the current English-dominated landscape. This opens the door for future work in building practical applications, from visually-grounded speech recognition to localized retrieval, for the hundreds of millions of people in developing countries.

\FloatBarrier

{
    \small
    \bibliographystyle{ieeenat_fullname}
    \bibliography{main}
}

\end{document}